\documentclass[lettersize,journal]{IEEEtran}
\usepackage{amsmath,amsfonts}
\usepackage{algorithmic}
\usepackage{algorithm}
\usepackage{array}
\usepackage[caption=false,font=normalsize,labelfont=sf,textfont=sf]{subfig}
\usepackage{textcomp}
\usepackage{stfloats}
\usepackage{url}
\usepackage{verbatim}
\usepackage{graphicx}
\usepackage{cite}
\usepackage{xcolor}
\usepackage{booktabs}
\usepackage{arydshln}
\begin{document}

\title{Rethinking Cross-Dose PET Denoising: Mitigating Averaging Effects via Residual Noise Learning}

\author{Yichao Liu, Zongru Shao, Yueyang Teng, Junwen Guo
        % <-this % stops a space
% <-this % stops a space
%Manuscript received April 19, 2021; revised August 16, 2021.
\thanks{ \textit{Corresponding author: Yueyang Teng and Junwen Guo}}

\thanks{Yichao Liu is with IWR, Heidelberg University, 69120, Heidelberg, Germany}
\thanks{Zongru Shao is with Silicon Austria Labs, 4040, Linz, Austria}
\thanks{Yueyang Teng is with College of Medicine and Biological Information Engineering, Key Laboratory of Intelligent Computing in Medical Image, Ministry of Education, Northeastern University, 110169, Shenyang, China (e-mail: tengyy@bime.neu.edu.cn)}
\thanks{Junwen Guo is with Department of Epidemiology \& Global Health, Umeå University, 90187, Umeå, Sweden (e-mail:junwen.guo@umu.se)}
}

\markboth{Journal of TRANSACTIONS ON RADIATION AND PLASMA MEDICAL SCIENCES,} %~Vol.~14, No.~8, August~2021}%
{Shell \MakeLowercase{\textit{et al.}}: A Sample Article Using IEEEtran.cls for IEEE Journals}


\maketitle

\begin{abstract}
%Experiments on large-scale multi-dose PET datasets from two medical centers demonstrate that the proposed method outperforms the “one-size-for-all” model, individual dose-specific U-Net models, and dose-conditioned approaches, achieving improved denoising performance. These results indicate that residual noise learning effectively mitigates the averaging effect and enhances generalization for cross-dose PET denoising.
Cross-dose denoising for low-dose positron emission tomography (LDPET) has been proposed to address the limited generalization of models trained at a single noise level. However, neural networks trained on a specific dose level often fail to generalize to other dose conditions due to variations in noise magnitude and statistical properties. Conventional “one-size-for-all” models attempt to mitigate this variability but tend to learn averaged representations across noise levels, resulting in degraded performance. In this work, we analyze this limitation and show that standard training formulations implicitly optimize an expectation over heterogeneous noise distributions, causing the network to learn an averaged denoising mapping that cannot accurately model dose-specific noise characteristics. We propose a unified residual noise learning framework that estimates noise directly from low-dose PET images rather than predicting full-dose images. Experiments on large-scale multi-dose PET datasets from two medical centers demonstrate that the proposed method outperforms the “one-size-for-all” model, individual dose-specific U-Net models, and dose-conditioned approaches, achieving improved denoising performance. These results indicate that residual noise learning effectively mitigates the averaging effect and enhances generalization for cross-dose PET denoising.
\end{abstract}


\section{Introduction}
\IEEEPARstart{P}{ositron} emission tomography (PET) plays a central role in oncologic staging \cite{Juweid2006PositronemissionTAA, Langer2010ASRA}, therapy response assessment \cite{Lloyd2010TheROA,Kitajima2017PresentAFA}, and longitudinal disease monitoring \cite{Liberini2021NSCLCBTA,Salem2022PETCTICA}. However, PET image quality is intrinsically constrained by limited photon counts, which are dictated by injected dose and acquisition time \cite{Vandenberghe2020StateOTA}. Dose reduction is clinically desirable to minimize radiation exposure, particularly in vulnerable populations and longitudinal studies, yet reduced counts induce severe Poisson noise, contrast degradation, and quantification bias in standardized uptake value (SUV) measurements \cite{Wang2020MachineLIA, Sanaat2021DeepTOFSinoADA, Yi2017SharpnessAwareLCA}. These effects compromise lesion detectability and downstream clinical decision-making.

Traditional method, such as iterative reconstruction algorithms and post-reconstruction filtering techniques, have historically proven their effectiveness in the low-dose PET images denoising \cite{novosad2016mr, golla2017partial,shepp1982maximum, karaoglanis2015iterative}. However, those methods are either time-consuming or prone to producing over-smoothed results that compromise fine structural details.

Recently, deep learning (DL) has emerged as a superior alternative to conventional iterative reconstruction and post-reconstruction filtering for PET image denoising \cite{hashimoto2022pet,spuhler2020full}. Convolutional neural networks (CNNs), in particular, have been extensively explored across several research directions. To improve feature representation, Xiang \textit{et al.} \cite{xiang2017deep} introduced a deep auto-context CNN that iteratively refines predictions to maintain image quality at lower radiation doses. Addressing resolution loss, Spuhler et al. \cite{spuhler2020full} developed a dilated CNN (dNet) that expands the network's receptive field without the need for downsampling. Beyond single-modality approaches, joint filtering frameworks have been proposed to denoise dynamic PET data by leveraging anatomical guidance from high-resolution MRI \cite{he2021dynamic}. More recently, research has shifted toward task-specific optimization, such as the LeqMod strategy, which utilizes downstream lesion quantification to ensure clinical accuracy in denoised images \cite{xia2025leqmod}. Beyond standard CNNs, Generative Adversarial Networks (GANs) have been extensively adapted for low-dose PET denoising to better preserve high-frequency details. For instance, Ouyang \textit{et al.} \cite{ouyang2019ultra} utilized feature matching and task-specific perceptual loss to synthesize standard-dose amyloid PET images from ultra-low-dose data (1\% counts). To enhance training stability and quantitative fidelity, Zhou \textit{et al.} \cite{zhou2020supervised} introduced CycleWGANs, which integrates cycle-consistency with the Wasserstein distance metric. More recently, the focus has shifted toward model robustness; Xue \textit{et al.} \cite{xue2022cross} developed a GAN-based approach for cross-scanner and cross-tracer recovery, demonstrating that a model trained on FDG-PET can successfully generalize to other tracers and hardware without retraining. Despite these advancements, they face a critical limitation: models trained on a specific noise level often fail to generalize well across diverse acquisition conditions. This performance degradation is caused by domain shift, the difference in joint probability between training and testing data, resulting in inaccurate estimations when there is a mismatch in count levels \cite{xie2023unified}. While training a "one-size-fits-all" model for cross-dose denoising may seem like a straightforward remedy, such models typically develop a bias toward the high-noise end of the spectrum, ultimately compromising their performance \cite{liu2022personalized}.

\begin{figure*}[h]
    \centering
    \includegraphics[width=\linewidth]{main_plot.png}
    \includegraphics[width=0.8\linewidth]{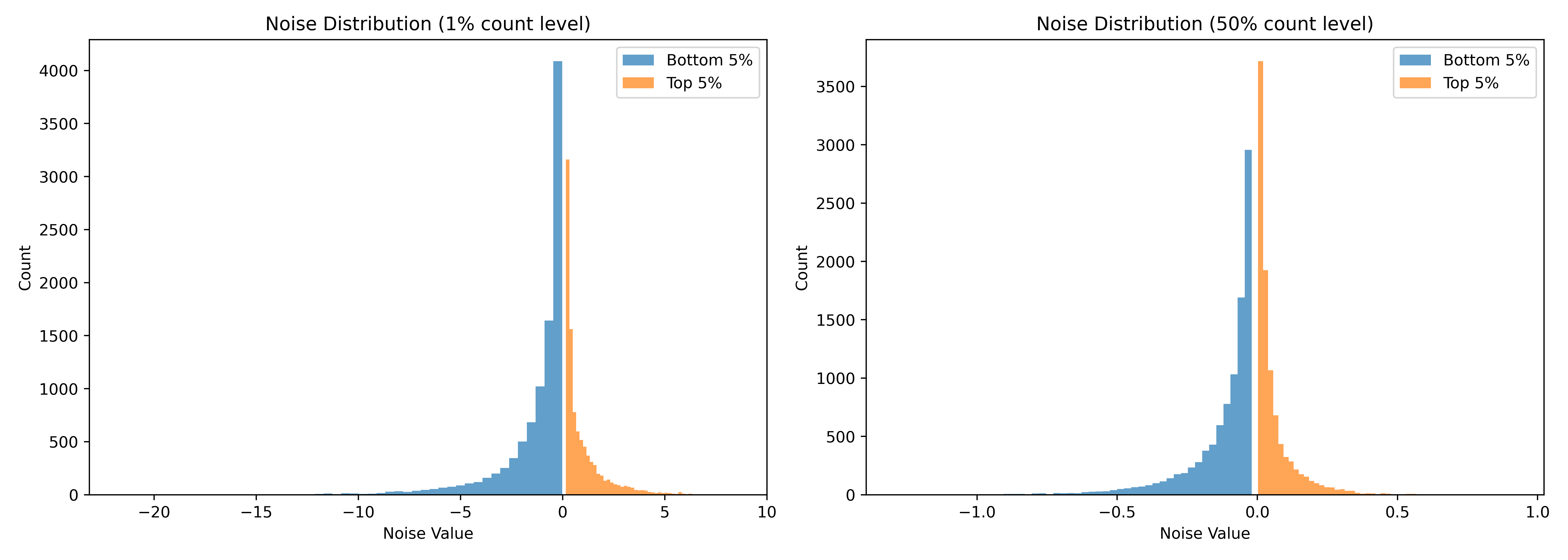}
    \caption{Noise analysis for 1/100 noise level and 1/2 noise level LDPET images. Top row: noise image is calculated by the difference between FDPET and LDPET. Bottom row: top and bottom 5\% noise pixel distribution.}
    \label{fig:noise}
\end{figure*}

\begin{figure*}[h]
    \centering
    \includegraphics[width=0.8\linewidth]{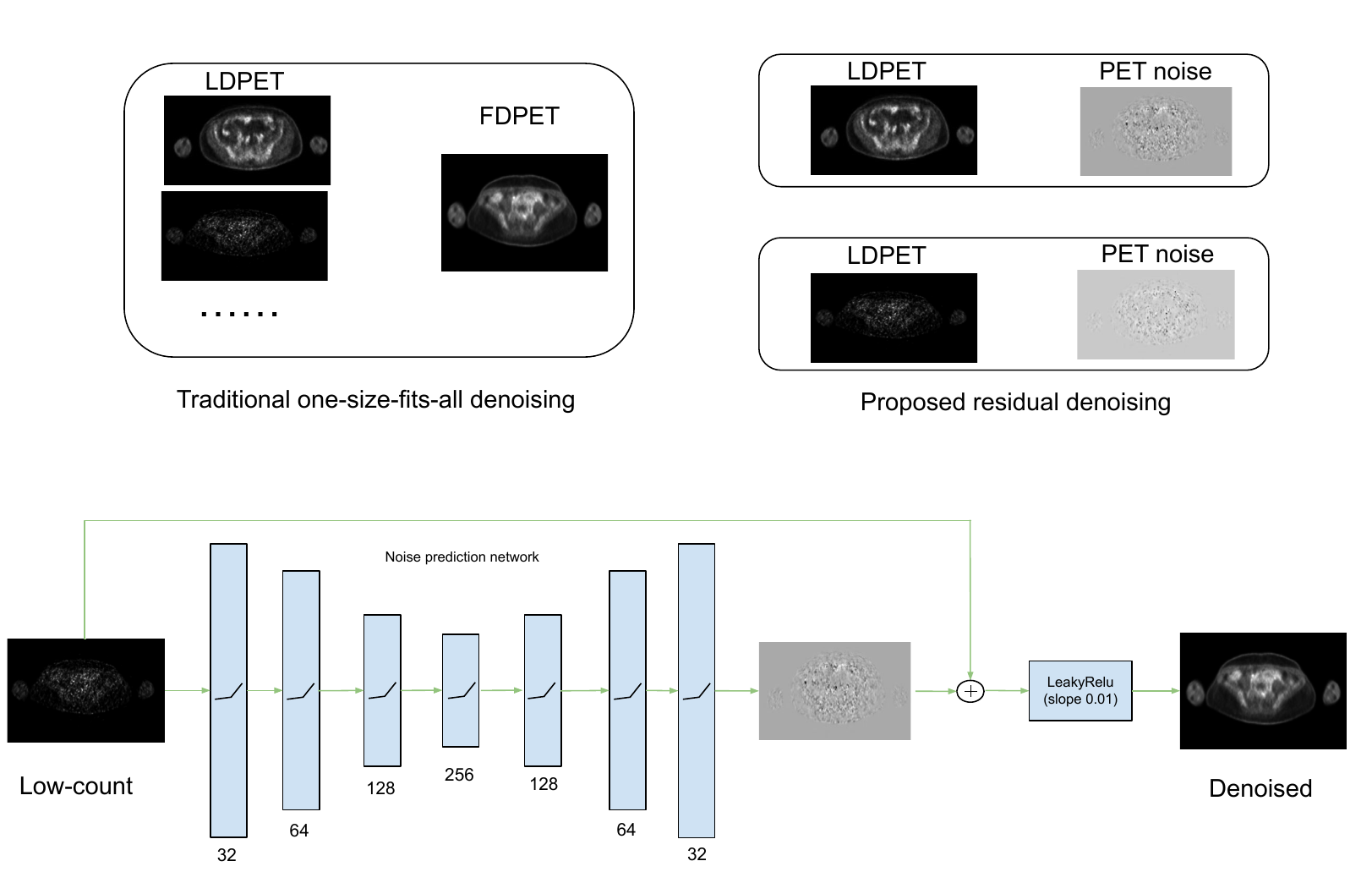}
    \caption{The overview of the proposed method. Top row presents the difference between the traditional on-size-fits-all denoising method and our proposed residual denoising method. Our method avoids learning an expectation across varying dose levels. The bottom row details the network architecture. The model takes a LDPET image as input and predicts the underlying noise. To prevent the loss of negative noise components, standard ReLU activations within the U-Net are replaced with LeakyReLU activation function. During training, the final denoised PET prediction is processed via a LeakyReLU with a shallow slope 0.01 to prevent neuron death.}
    \label{fig:overview}
\end{figure*}

To address the challenges of varying count statistics, unified models have been proposed to facilitate cross-dose level denoising. The Unified Noise-Aware Network (UNN) was introduced to adaptively merge multiple specialized denoising subnetworks through a learned weighting mechanism, achieving robust PET image quality across diverse and unknown photon count levels \cite{xie2023unified}. Building on this framework, the ST-UNN architecture replaces traditional convolutional layers with Swin Transformer blocks and increases subnetwork granularity to eight levels, specifically targeting the complexities of ultra-low-dose PET imaging (below 10\% count) \cite{azimi2025deep}. However, the UNN-based approach remains computationally demanding, as it requires the pretraining, loading, and storage of multiple specialized subnetworks during training. To improve efficiency, an alternative approach utilizing Continuous Adversarial Domain Generalization (CADG) employs a single model with a continuous discriminator in the feature space \cite{liu2024cross}. This framework enforces the extraction of noise-level invariant features, enabling the network to generalize effectively across arbitrary and unseen PET count levels. A potential drawback of such domain generalization is that enforced domain invariance can inadvertently degrade performance in scenarios where the domain and label are entangled \cite{guan2021domain,zhou2022domain}. While the authors of the CADG study argue that PET denoising primarily involves covariate shift, where the full-count label distribution remains consistent regardless of the noise level, this assumption may not hold in broader clinical practice. Nevertheless, to the best of our knowledge, no existing studies have explicitly identified that cross-dose training essentially forces the model to minimize an expectation loss across varying noise distributions.

In this work, we show that conventional cross-dose PET denoising suffers from optimization target mismatch caused by heterogeneous conditional noise distributions. Existing unified models reduce this issue through architectural specialization, but still optimize toward averaged reconstruction targets. We reformulate the problem as conditional residual noise estimation. Second, we propose a solution centered on residual noise learning; as established by DnCNN \cite{zhang2017beyond}, this architecture facilitates faster convergence and superior training stability by focusing the network’s learning capacity on the noise distribution. Third, we validate our approach through experiments on two clinical datasets, demonstrating that our method achieves superior performance and high robustness across diverse patient cohorts.

\section{Methods}
\subsection{Averaging effect of supervised cross-dose denoising}
%\subsection{Supervised denoising}
In PET denoising, given a low-dose image $\mathbf{x} \in \mathcal{R}^{H \times W}$, the corresponding full-dose image $\mathbf{y} \in \mathcal{R}^{H \times W}$, where $W$ and $H$ are width and height, and $f_{\theta}$ is denoising network, the denoising process can be formulated as:
\begin{equation}
    arg \, \underset{\theta}{min} \sum_{i}L(f_{\theta}(\mathbf{x}_{i}),\mathbf{y}_{i})
    \label{supervised}
\end{equation}
Here, $L$ is the loss function. $\theta$ is the network parameter. In the case of "one-size-fits-all" model, Under a 'one-size-fits-all' model, we assume a specific probability of occurrence for the patient-dose pair $(\mathbf{x}_{i=l,j},\mathbf{y}_{i=l,j})$ at patient $l$, dose level $j$, Consequently, Eq. \ref{supervised} can be statistically rewritten as:
\begin{align}
    \theta^{*}&=arg \, \underset{\theta}{min} \sum_{j=1}p(\mathbf{x}_{i=l,j},\mathbf{y}_{i=l})L(f_{\theta}(\mathbf{x}_{i=l,j}),\mathbf{y}_{i=l}) \nonumber \\
    &=arg \, \underset{\theta}{min} \mathbb{E}_{(\mathbf{x}_{i=l},\mathbf{y}_{i=l})}[L(f_{\theta}(\mathbf{x}_{i=l}),\mathbf{y}_{i=l})]
    \label{exp1}
\end{align}
Here, the paired PET images $(\mathbf{x}, \mathbf{y})$ are related by $\mathbf{y} = \mathbf{x} + \mathbf{n}_0$, where $\mathbf{n}_0$ represents the observed random noise. It follows that the joint probability can be decomposed as $p(\mathbf{x}_{i=l,j},\mathbf{y}_{i=l})=p(\mathbf{x}_{i=l,j})p(\mathbf{y}_{i=l}|\mathbf{x}_{i=l,j})$. Consequently, Eq. \ref{exp1} is equivalent to:

\begin{align}
    \theta^{*}&=arg \, \underset{\theta}{min}\sum_{j=1}p(\mathbf{x}_{i=l,j})p(\mathbf{y}_{i=l}|\mathbf{x}_{i=l,j})L(f_{\theta}(\mathbf{x}_{i=l,j}),\mathbf{y}_{i=l}) \nonumber \\
    &=arg \, \underset{\theta}{min} \mathbb{E}_{\mathbf{x}_{i=l}}[\mathbb{E}_{\mathbf{y}_{i=l}|\mathbf{x}_{i=l}}[L(f_{\theta}(\mathbf{x}_{i=l}),\mathbf{y}_{i=l})]]
    \label{all-in-one}
\end{align}
Under common regression losses such as mean squared error (MSE) or mean absolute error (MAE), the optimal estimator learned by the network converges toward the conditional expectation of the target given the input:

$f^{*}(x)=\mathbb{E}[y|x]$

In cross-dose PET denoising, however, the conditional distribution
$p(y|x)$ is influenced by heterogeneous count statistics originating from different dose levels, scanners, and acquisition conditions. Consequently, the learned estimator implicitly minimizes an expectation over multiple conditional noise distributions rather than a dose-specific target distribution. This causes the reconstructed output to converge toward an averaged solution across heterogeneous noise domains, resulting in over-smoothed predictions and degraded restoration fidelity. We refer to this phenomenon as the averaging effect in cross-dose PET denoising. This approach is sub-optimal compared to a dose-conditioned network, a distinction we explore in detail through our experimental results.

In contrast to the 'one-size-fits-all' model, the dose-conditioned network embeds dose level information alongside the LDPET image as a joint input. For a given dose level $m$, the probability dependence is expressed as $p(\mathbf{x}_{i=l,j},\mathbf{m}_j,\mathbf{y}_{i=l})=p(\mathbf{x}_{i=l,j},\mathbf{m}_j)p(\mathbf{y}_{i=l}|\mathbf{x}_{i=l,j},\mathbf{m}_j)$. Consequently, Eq. \ref{all-in-one} can be rewritten as:
\begin{align}
    \theta^{*}&=arg \, \underset{\theta}{min}\sum_{j=1}p(\mathbf{x}_{i=l,j},\mathbf{m}_j)p(\mathbf{y}_{i=l}|\mathbf{x}_{i=l,j},\mathbf{m}_j) \nonumber \\
    &L(f_{\theta}(\mathbf{x}_{i=l,j},\mathbf{m}_j),\mathbf{y}_{i=l}) \nonumber \\
    &=arg \, \underset{\theta}{min} \mathbb{E}_{\mathbf{x}_{i=l},\mathbf{m}_j}[\mathbb{E}_{\mathbf{y}_{i=l}|\mathbf{x}_{i=l},\mathbf{m}_j}[L(f_{\theta}(\mathbf{x}_{i=l},\mathbf{m}_j),\mathbf{y}_{i=l})]]
    \label{dose_embed}
\end{align}
In this formulation, the expectation is calculated specifically for each dose level. Consequently, the model learns a dose-dependent denoising mapping optimized for that particular noise distribution, requiring count level information during inference. Although UNN \cite{xie2023unified} and ST-UNN \cite{azimi2025deep} adopt a multi-expert framework by leveraging multiple dose-specific models, they do not fundamentally eliminate the cross-dose averaging effect.  
\subsection{Asymmetric Residual Degradation Analysis}
In conventional natural image residual denoising methods such as DnCNN, stable performance is typically achievable because the additive noise is assumed to follow a symmetric Gaussian distribution. In contrast, PET denoising is governed by Poisson statistics, which introduce inherent asymmetry in the noise distribution. As illustrated in Fig. \ref{fig:noise}, LDPET images exhibit a strongly skewed pattern. In particular, the maximum intensity values can become higher than those in normal dose PET, with this effect being especially pronounced at the 1\% count level. The bottom row further analyzes the top and bottom 5\% of noise residuals, revealing a clear bias toward negative values. This asymmetry can bias the learning objective toward negative residuals. Consequently, a substantial number of reconstructed pixels may even become negative, which is physically implausible for PET activity distributions.

\begin{table*}[!ht]
\caption{Quantitative performance comparison across two datasets. The best results for each count level are highlighted in bold. Our proposed method consistently outperforms baseline approaches across the majority of count levels. Symbols $\dagger$ and $*$ denote statistical significance at $p < 0.005$ and $p < 0.05$, respectively, relative to individual denoising U-Net models at corresponding noise levels. }
\label{tab:comparison}
\centering
\begin{tabular}{ccccccc}
\toprule
\multicolumn{7}{c}{Dataset from University of Bern (Siemens biograph vision quara scanner)} \\
\midrule
PSNR & 1\% count level& 2\% count level&5\% count level& 10\% count level& 25\% count level&50\% count level\\
\hdashline
LDPET&25.201&30.083	&35.090&	38.292&	42.897&	47.485\\
Individual Unet models&34.083&35.852&	37.454&	40.076&	43.664&	46.759\\
one Unet for all& 31.538&	32.616&	33.605&	34.050&	34.456&	34.596\\
UNN weighted sum&33.535&36.073&	38.361&	39.467&	40.462& 40.820\\
Unet with dose embedding& \textbf{34.825}&	\textbf{37.127}&	39.658&	41.334&	\textbf{44.021}&	\textbf{46.830}\\
Proposed & 34.580*&	37.034*&	\textbf{39.719\textdagger}&	\textbf{41.508*}&	43.929&	45.860\\
\midrule
SSIM & 1\% count level& 2\% count level&5\% count level& 10\% count level& 25\% count level&50\% count level\\
\hdashline
LDPET&0.910&0.929&0.953&0.966&0.982&0.992\\
Individual Unet models&0.960&	0.967&	0.972&	0.981&	0.988&	0.993\\
one Unet for all&0.500	&0.504&	0.510&	0.514&	0.518&	0.520\\
UNN weighted sum&0.953&	0.966&	0.977&	0.981&	0.986& 0.988\\
Unet with dose embedding&\textbf{0.961}&	0.971&	0.980&	0.984&	0.989&	0.993\\
Proposed&0.959&	\textbf{0.971*}&	\textbf{0.980\textdagger}&	\textbf{0.984*}&	\textbf{0.990}&	\textbf{0.993}\\
\midrule
RMSE & 1\% count level& 2\% count level&5\% count level& 10\% count level& 25\% count level&50\% count level\\
\hdashline
LDPET &0.353&	0.202&	0.115&	0.079&	0.047&	0.028\\
Individual Unet models&0.141&	0.113&	0.097&	0.070&	0.046&	0.033\\
one Unet for all&0.195&	0.171&	0.156&	0.150&	0.146&	0.145\\
UNN weighted sum&0.143&	0.107&	0.085&	0.076&	0.070& 0.068\\
Unet with dose embedding&\textbf{0.124}&	0.094&	0.072&	0.060&	\textbf{0.044}&	\textbf{0.033}\\
Proposed&0.127*&	\textbf{0.094*}&	\textbf{0.070\textdagger}&	\textbf{0.058*}&	0.045&	0.038\\
\bottomrule
\end{tabular}
\vspace{1cm} \\
\begin{tabular}{ccccccc}
\toprule
\multicolumn{7}{c}{Dataset from Shanghai Ruijin Hospital(United Imaging uExplorer scanner)} \\
\hline
PSNR & 1\% count level& 2\% count level&5\% count level& 10\% count level& 25\% count level&50\% count level\\
\hdashline
LDPET&23.049&	27.735&	32.396&	35.373&	39.759&	44.274\\
Individual Unet models&31.874&	33.570&	36.269&	37.789&	40.756&	43.394\\
UNN weighted sum&31.247&	33.822&	35.959&	36.952&	37.848& 38.239\\
Unet with dose embedding& 32.079&	34.070&	36.234&	37.907&	40.565&	43.699\\
Proposed & \textbf{32.458\textdagger}&	\textbf{34.667\textdagger}&	\textbf{37.065\textdagger}&	\textbf{38.786\textdagger}&	\textbf{41.561*}&	\textbf{44.381*}\\
\hline
SSIM & 1\% count level& 2\% count level&5\% count level& 10\% count level& 25\% count level&50\% count level\\
\hdashline
LDPET&0.913&	0.933&	0.958&	0.971&	0.986&	0.994\\
Individual Unet models&0.955&	0.965&	0.976&	0.982&	0.989&	0.994\\
UNN weighted sum&0.949&	0.965&	0.976&	0.980&	0.983& 0.98\\
Unet with dose embedding&0.957&	0.969&	0.978&	0.983&	0.990&	0.994\\
Proposed&\textbf{0.963\textdagger}&	\textbf{0.973\textdagger}&	\textbf{0.981\textdagger}&	\textbf{0.985\textdagger}&	\textbf{0.991\textdagger}&	\textbf{0.994*}\\
\hline
RMSE & 1\% count level& 2\% count level&5\% count level& 10\% count level& 25\% count level&50\% count level\\
\hdashline
LDPET&0.247&	0.142&	0.083&	0.059&	0.036&	0.022\\
Individual Unet models&0.098&	0.079&	0.057&	0.047&	0.034&	0.026\\
UNN weighted sum&0.101&	0.075&	0.060&	0.055&	0.050& 0.049\\
Unet with dose embedding&0.093&	0.073&	0.057&	0.047&	0.035&	0.024\\
Proposed&\textbf{0.089*}&\textbf{0.067\textdagger}&	\textbf{0.050\textdagger}&	\textbf{0.041\textdagger}&	\textbf{0.031\textdagger}&	\textbf{0.023*}\\
\bottomrule
\end{tabular}
\end{table*}

\begin{figure*}[h]
    \centering
    \includegraphics[width=0.45\linewidth]{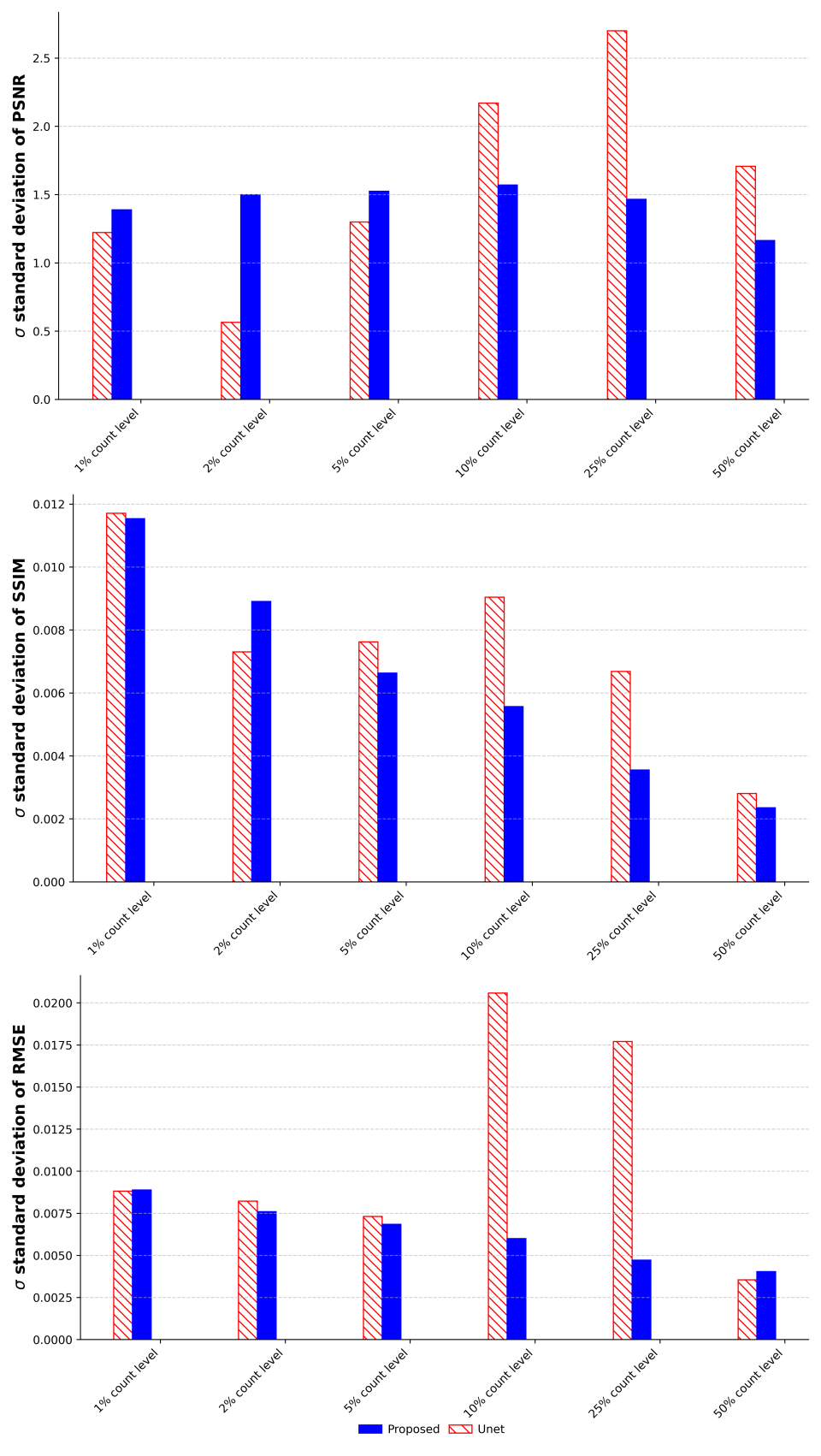}
    \includegraphics[width=0.45\linewidth]{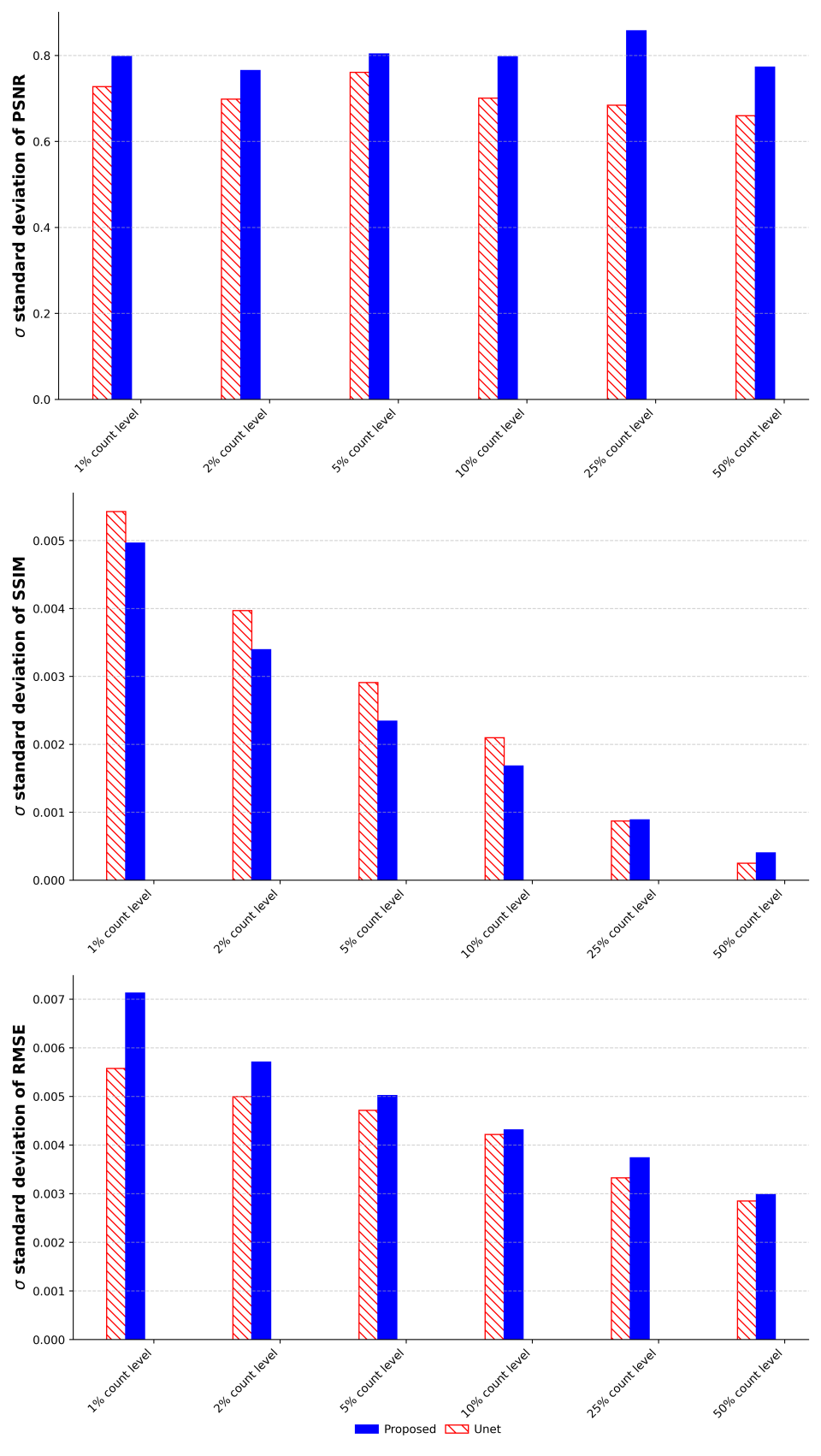}
    \caption{Quantitative comparison of PSNR, SSIM and RMSE of standard deviation on two datasets, University of Bern (left) and Shanghai Ruijin hospital (right). The blue bar is our proposed residual noise learning method, and the red bar is the individual trained Unet for each dose level.}
    \label{fig:std}
\end{figure*}

\begin{figure*}[h]
    \centering
    \includegraphics[width=\linewidth]{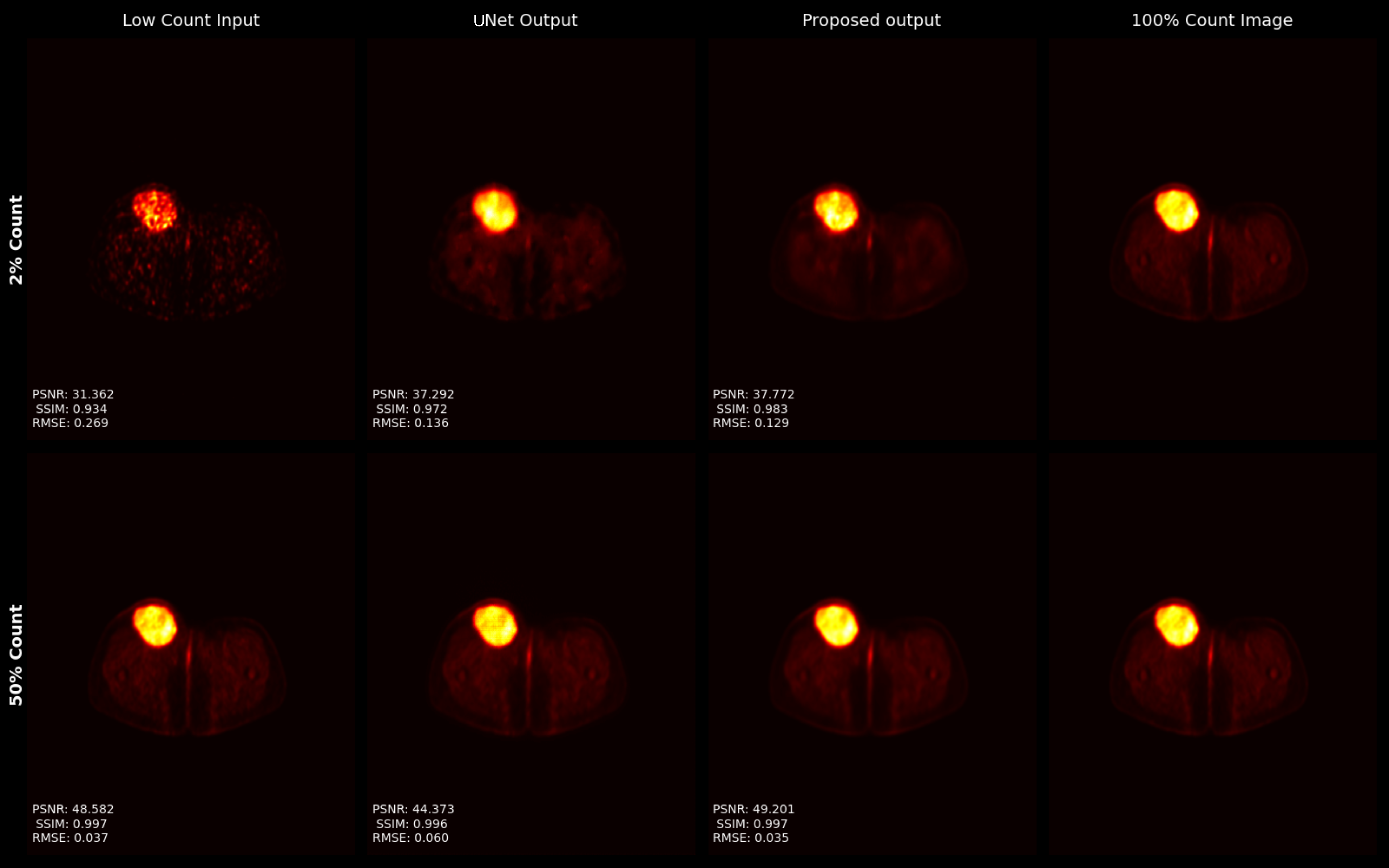}
    \caption{Qualitative comparison of denoised upper abdomen using different methods for 1/50 noise level and 1/2 noise level LDPET images. The sample is selected from dataset acquired by Shanghai Ruijin Hospital.}
    \label{fig:shanghai_reginal}
\end{figure*}
\begin{figure}
    \centering
    \includegraphics[width=\linewidth]{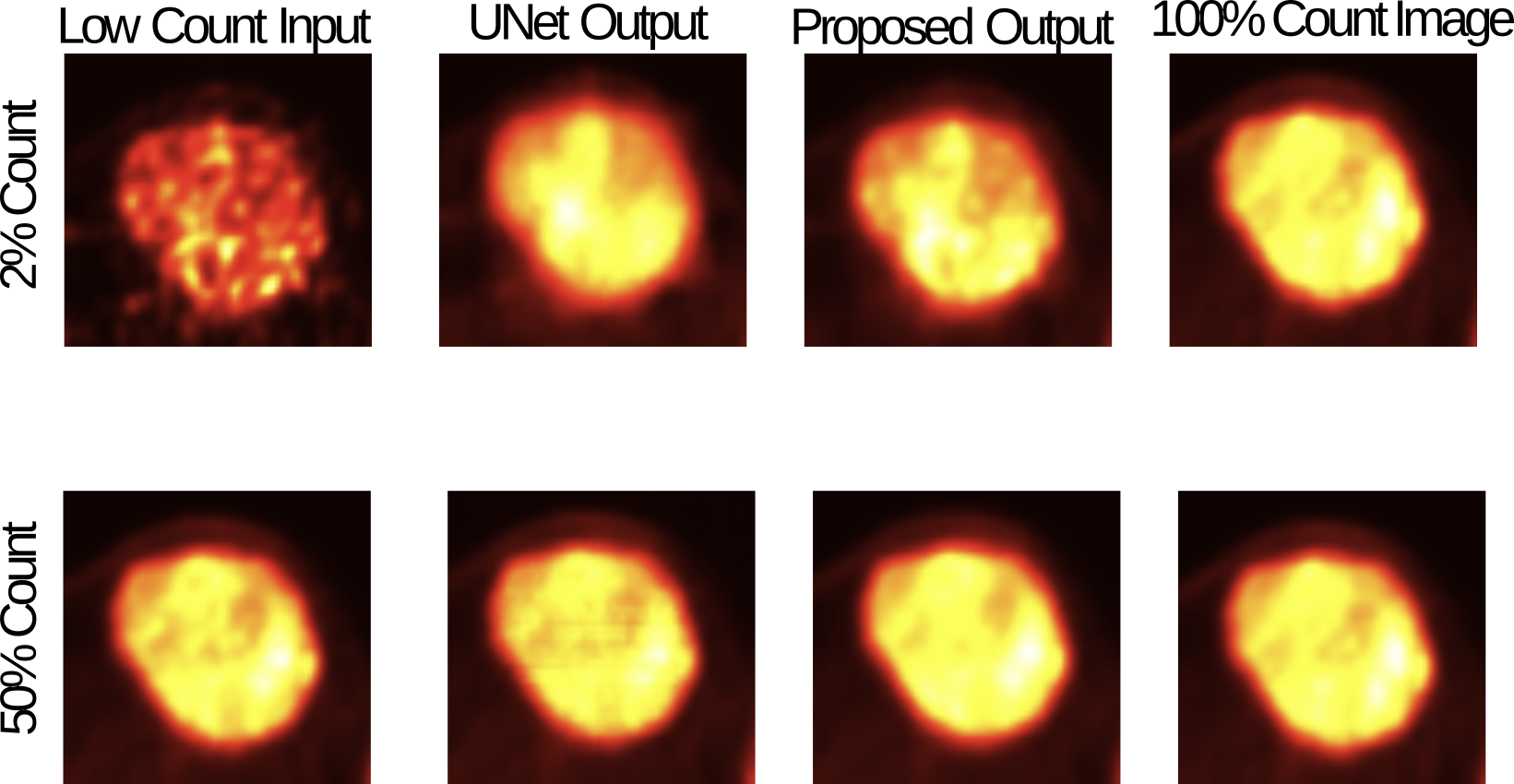}
    \caption{Zoom in details of upper abdomen comparison.}
    \label{fig:zoom_in_S}
\end{figure}

\begin{figure*}[h]
    \centering
    \includegraphics[width=\linewidth]{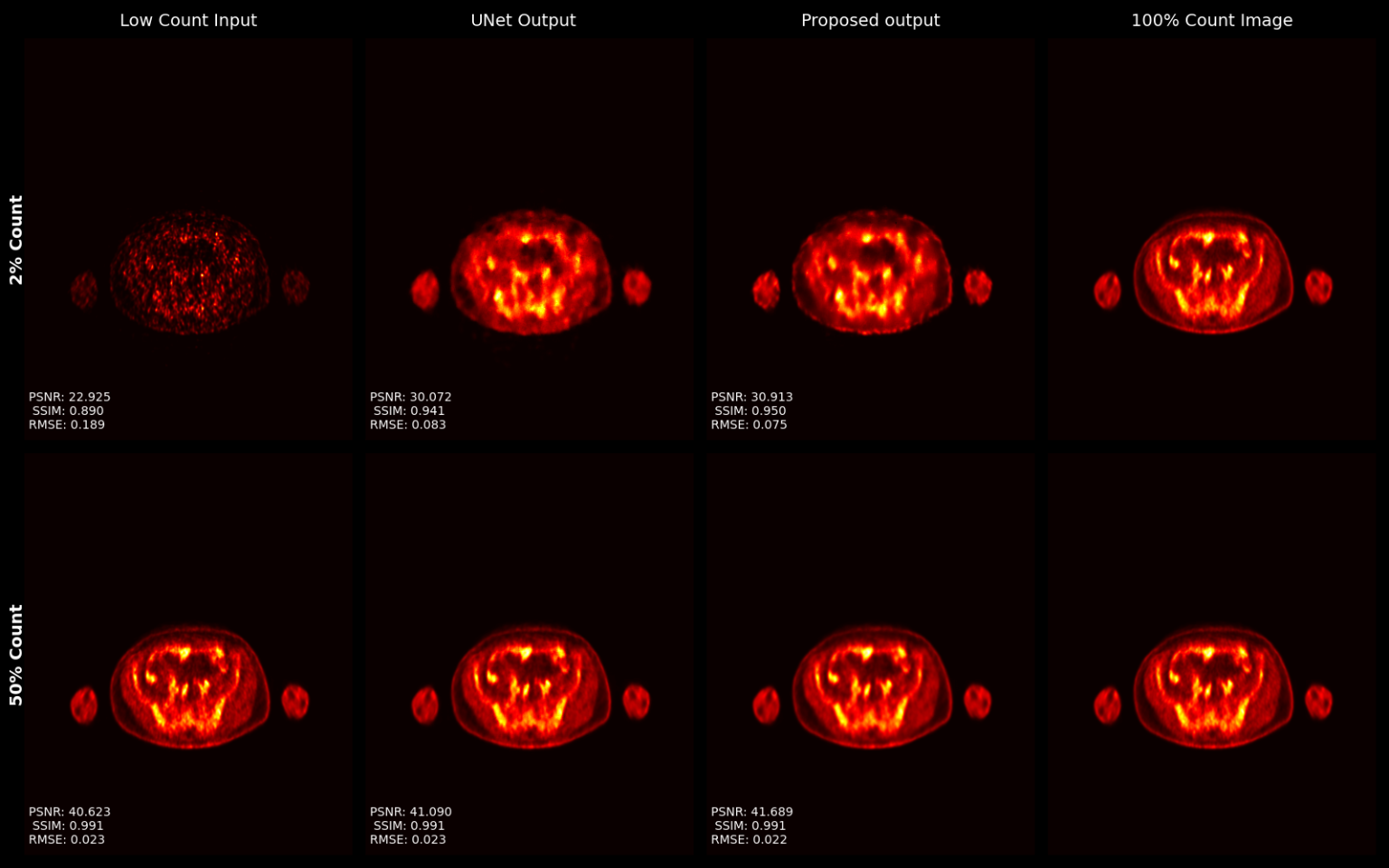}
    \caption{Qualitative comparison of denoised lower abdomen using different methods for 1/50 noise level and 1/2 noise level LDPET images. The sample is selected from dataset acquired by the University of Bern.}
    \label{fig:bern_reginal}
\end{figure*}

\begin{figure}
    \centering
    \includegraphics[width=\linewidth]{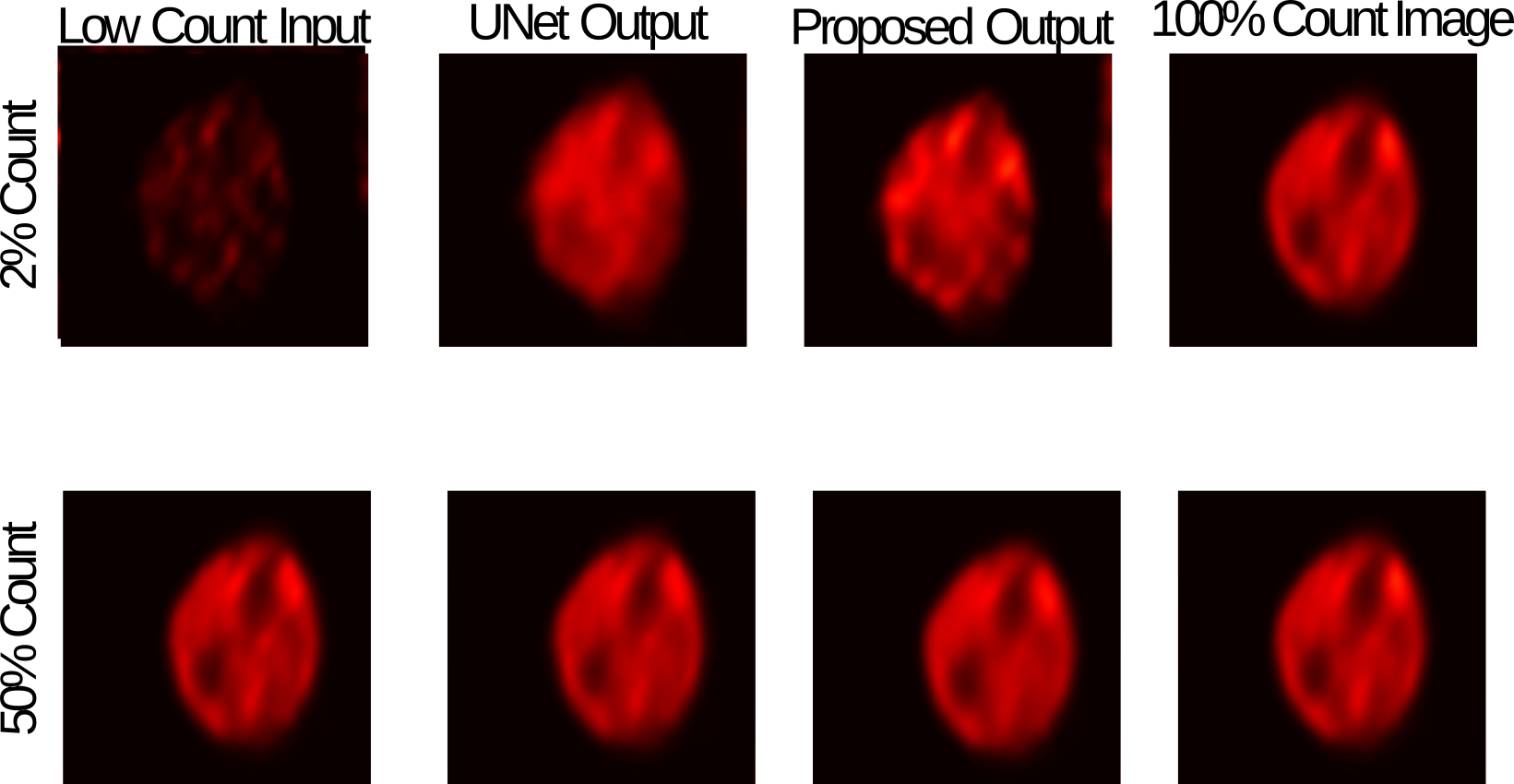}
    \caption{Zoom in details of lower abdomen comparison.}
    \label{fig:zoom_in_B}
\end{figure}
\begin{figure*}[h]
    \centering
    \includegraphics[width=\linewidth]{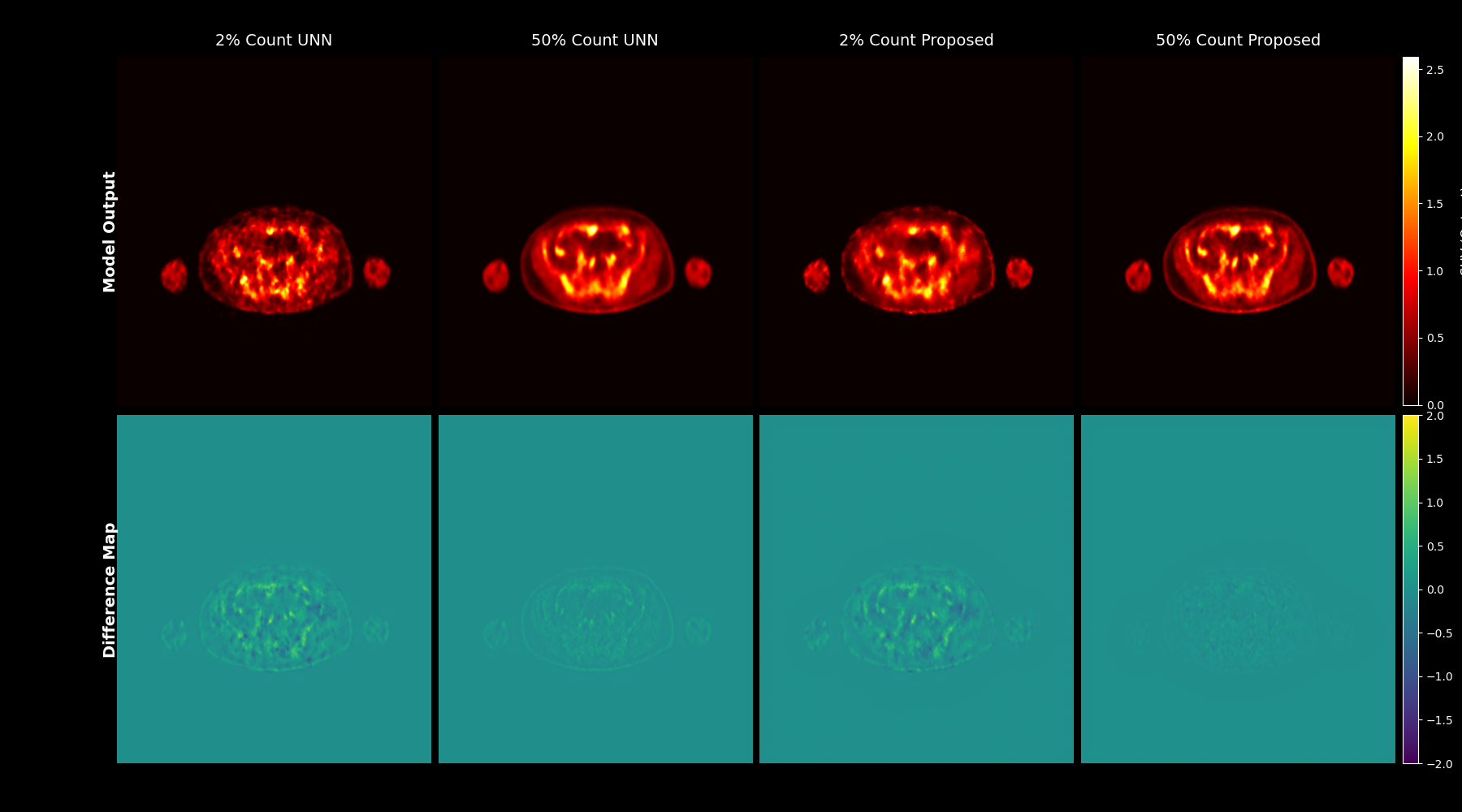}
    \caption{Comparison of denoised lower abdomen using different methods for 1/50 noise level and 1/2 noise level LDPET images. The sample is selected from dataset acquired by the University of Bern. Top row is denoised PET image by UNN and proposed method. Bottom row is the difference between the denoised PET image and FDPET. }
    \label{fig:bern_reginal_diff}
\end{figure*}

\subsection{Residual denoising}
To circumvent these limitations, we approach the problem from a different perspective. Rather than directly predicting the FDPET image, the model is tasked with estimating the specific noise associated with each dose level. By defining the noise component as $\mathbf{n}_j = \mathbf{y} - \mathbf{x}_j$, we represent each patient-dose pair as ($\mathbf{x}_{i=l,j},\mathbf{n}_{i=l,j}$). . Consequently, Eq. \ref{all-in-one} is reformulated as follows:

\begin{align}
    \theta^{*}&=arg \, \underset{\theta}{min}\sum_{j=1}p(\mathbf{x}_{i=l,j})p(\mathbf{n}_{i=l,j}|\mathbf{x}_{i=l,j})L(f_{\theta}(\mathbf{x}_{i=l,j}),\mathbf{n}_{i=l,j}) \nonumber \\
    &=arg \, \underset{\theta}{min} \mathbb{E}_{\mathbf{x}_{i=l}}[\mathbb{E}_{\mathbf{n}_{i=l,j}|\mathbf{x}_{i=l}}[L(f_{\theta}(\mathbf{x}_{i=l}),\mathbf{n}_{i=l,j})]]
\end{align}
Here, the expectation of the noise $\mathbf{n}_{i=l,j}$ is conditioned strictly on the LDPET image $\mathbf{x}_{i=l,j}$ at its corresponding dose level. While residual noise learning does not eliminate the expectation formulation itself, it changes the optimization target from anatomically complex PET reconstructions to residual noise distributions with substantially lower structural variability. Since the residual component is more directly associated with dose-dependent degradation patterns, the network can more effectively exploit implicit conditioning through the input $x$, thereby reducing cross-dose interference during optimization.

As DnCNN \cite{zhang2017beyond} demonstrates stable training and ease of optimization, we adapt it for our cross-noise-level denoising task. Based on the supervised learning formulation in Eq.~\ref{supervised}, the denoising process can be expressed as follows:
\begin{equation}
    arg \, \underset{\theta}{min} \sum_{i}L(f_{\theta}(x_{i})+x_{i},y_{i})
    \label{MAE}
\end{equation}
Note that the output of $f_\theta$ corresponds to the predicted PET noise $\mathbf{n}_{i=l,j}$. However, our empirical observations indicate that directly applying DnCNN to PET denoising is suboptimal. This limitation arises because the residual noise, defined as $y_i-x_i$, can take negative values, as discussed in the last sub-section and shown in Fig. \ref{fig:noise}, whereas the  Rectified linear unit (ReLU) activation used in DnCNN suppresses all negative responses. To address this issue, we replace the ReLU activation in $f_{\theta}$ with LeakyReLU, which permits negative outputs and thus enables more accurate modeling of the noise distribution. Furthermore, a hard constraint is applied to the residual output, $f_{\theta}(x_{i}) + x_{i}$, using a LeakyReLU activation function with a shallow slope of 0.01. This choice ensures the output remains strictly positive while preventing the 'dying ReLU' problem, where neurons become permanently inactive during training. The overview is shown in the Fig. \ref{fig:overview}.

The Structural Similarity Index (SSIM) evaluates the structural correspondence between two images, yielding a maximum value of 1 for identical inputs. For this implementation, a $11 \times 11$ convolutional window was employed to compute the index. The SSIM is defined by the following expression:
\begin{align}
    SSIM(x,y)=\frac{(2\mu_x\mu_y+c_1)(\sigma_{xy}+c_2)}{(\mu^{2}_{x}+\mu^{2}_{y}+c_1)(\sigma^2_{x}+\sigma^2_{y}+c_2)}
\end{align}
where, $\mu_x$ and $\mu_y$ denote the local means of images $x$ and $y$, while $\sigma_x^2$ and $\sigma_y^2$ represent their respective variances. The term $\sigma_{xy}$ signifies the cross-covariance between the two images, and the constants $c_1$ and $c_2$ serve as stability parameters to prevent division by zero. These hyperparameters are defined as $c_1 = (k_1 L)^2$ and $c_2 = (k_2 L)^2$, where $L$ represents the dynamic range of pixel values, with the default values for $k_1$ and $k_2$ set to 0.01 and 0.03, respectively. SSIM loss is used to optimize the network parameters and is expressed as: $L_{SSIM} = 1 - \text{SSIM}(Y, X)$. The whole loss function for training can be expressed as 
\begin{equation}
    L=L_{MAE}+\lambda L_{SSIM}
\end{equation}
where $\mathcal{L}_{\text{MAE}}$ represents the Mean Absolute Error loss as defined in Eq. \ref{MAE}, and $\lambda$ serves as a regularization hyperparameter that balances the contribution of the reconstruction term against the SSIM loss. Based on the trial and error, we set $\lambda$ to 0.5.

\section{Experiments}
\subsection{Dataset}
We have evaluated our method on two multi-dose PET datasets. The datasets are from University of Bern, Dept. of Nuclear Medicine and School of Medicine \cite{xue2022cross}. The first dataset includes 209 subjects imaged with the $^{18}$F-FDG tracer. All data were acquired using a Siemens Biograph Vision Quadra whole-body PET/CT system. Images were reconstructed using the Ordered Subset Expectation Maximization (OSEM) algorithm with six iterations and five subsets, followed by the application of a 5-mm FWHM Gaussian filter. The reconstruction matrix size is $644 \times 440 \times 440$ with an isotropic voxel size of 1.65 mm$^3$.

The second dataset is from Ruijin Hospital, Shanghai, China \cite{xue2022cross}. This dataset includes $^{18}$F-FDG PET scans from 320 subjects, all acquired on a United Imaging Healthcare uEXPLORER total-body PET/CT system. Images were reconstructed using the OSEM algorithm (4 iterations, 20 subsets) followed by a 5-mm FWHM Gaussian post-filter. The resulting reconstruction matrix was $673 \times 360 \times 360$ with an anisotropic voxel size of $2.89 \times 1.67 \times 1.67 \text{ mm}^3$. 

The raw PET data were initially acquired in list-mode format, recording individual coincidence events chronologically. events were randomly selected from the total event pool to synthesize low-count datasets at 1\%, 2\%, 5\%, 10\%, 25\%, and 50\% of the original count levels. This approach ensures that the simulated data maintains the physical Poisson statistics inherent in PET acquisition, providing a realistic foundation for training our noise-aware framework.

For the training set, 9,000 samples were randomly extracted from 18 patients, while 6,000 samples were selected from 6 patients for testing. Data preprocessing involved converting raw pixel intensities into Standardized Uptake Values (SUVs), with the resulting values clipped to a fixed range.

\subsection{Implementation details}
We employ the U-Net architecture for our model. The network is trained using the Stochastic Gradient Descent (SGD) optimizer, configured with a momentum of 0.9, a learning rate of 0.01, and a weight decay of $1e-4$ \cite{robbins1951stochastic}. Following the implementation of Swin-Unet \cite{cao2022swin}, we apply a learning rate decay strategy to optimize convergence.
\begin{equation}
	lr=lr_{base}\cdot (1-\frac{n_{iter}}{M_{total}})^{\gamma}
\end{equation}
where $lr_{base}=0.01$, $n_{iter}$ is the current iterations, $M_{total}$ is the total iterations, $\gamma=0.85$ is decay factor. The model is trained for 80 epochs with a mini-batch size of 16. During inference, the batch size is set to 1. To ensure a fair evaluation, all baseline comparisons utilize the same training strategy. We evaluate our method against a standard U-Net and a modified U-Net that incorporates an additional embedded dose level as an input feature, which trained under the same conditions. 

For quantitative evaluation, evaluated model performance using three standard metrics following the implementation of Chen \textit{et al.} \cite{chen2017low}: peak signal-to-noise ratio (PSNR), SSIM, and root mean squared error (RMSE).

\subsection{Results}

\subsubsection{Quantitative comparison}

To evaluate the efficacy of our proposed residual noise learning network, we compared its performance against individual Unets trained on specific noise levels (ranging from 1/100 to 1/2 dose). Quantitative results are detailed in Table \ref{tab:comparison} and Fig. \ref{fig:std}. As shown in Table \ref{tab:comparison}, the denoising performance of standard models generally degrades as the dose level increases; at the 50\% dose level, where the input closely resembles the full-dose image, standard denoising models actually underperform compared to the raw LDPET in terms of PSNR. Our proposed residual noise learning network consistently outperforms both individual and dose-embedded Unets in most scenarios, with particularly significant gains observed in the Shanghai Ruijin Hospital dataset. We also evaluated a "one-size-fits-all" baseline, which exhibited a severe drop in both PSNR and SSIM. The poor SSIM results suggest that a naive unified model merely learns an averaged structure across dose levels, validating our analysis of the averaging effects in standard loss functions. Notably, while dose-embedded models require explicit dose information during inference, our method achieves superior or comparable results without requiring any prior dose metadata. Furthermore, at the 50\% dose level, our model successfully outperforms the raw LDPET PSNR for Ruijin Hospital data.

Regarding cohort stability, Fig. \ref{fig:std} illustrates the standard deviation of metrics across patients. In the University of Bern dataset, our method demonstrates superior stability across all noise levels. In contrast, individually trained U-Nets exhibit a significant spike in variance at 10\% and 25\% dose levels. This instability at intermediate doses stems from the inability of fixed kernels to adapt to volatile signal-to-noise ratios influenced by patient-specific factors, such as anatomy and BMI. In the second dataset, our method maintains stability comparable to individual models while demonstrating greater overall robustness across the patient population.

\begin{figure*}[h]
    \centering
    \includegraphics[width=\linewidth]{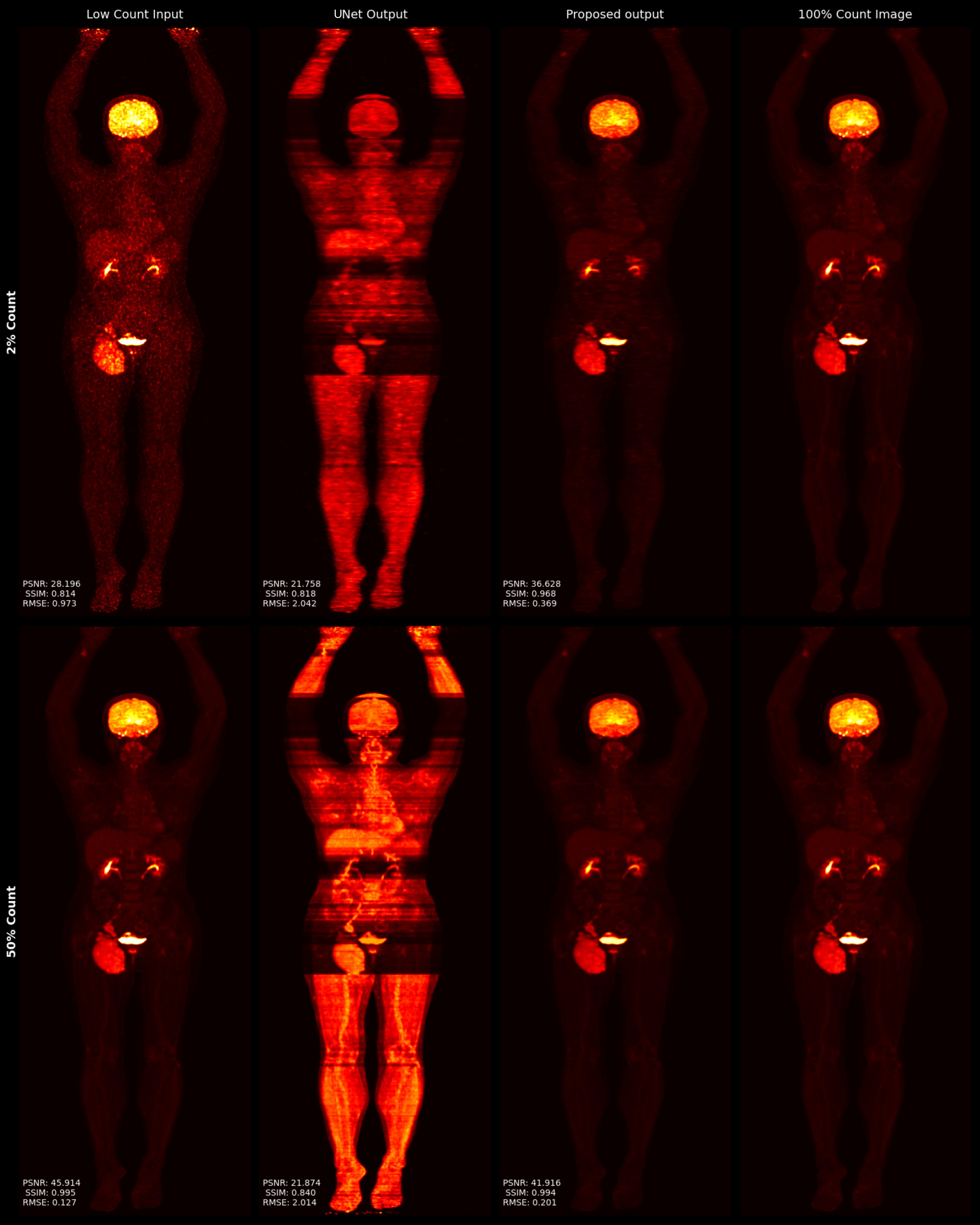}
    \caption{Qualitative comparison of whole-body PET denoising using different methods at $1/50$ (2\%) and $1/2$ (50\%) dose levels. Images are presented as Maximum Intensity Projections (MIP) along the longitudinal ($z$) axis. The rows display sample cases from the Shanghai Ruijin Hospital dataset. Columns from left to right: LDPET input, outputs from individually trained Unet models, results from the proposed network, and the FDPET ground truth.}
    \label{fig:whole_body}
\end{figure*}

\subsubsection{Qualitative comparison}
We provide qualitative comparisons for two sample cases: an upper abdomen scan from Shanghai Ruijin Hospital (Fig. \ref{fig:shanghai_reginal} and selected area in Fig. \ref{fig:zoom_in_S}) and a lower abdomen scan from the University of Bern (Fig. \ref{fig:bern_reginal} and selected area in Fig. \ref{fig:zoom_in_B}). As shown in Fig. \ref{fig:shanghai_reginal}, the 2\% dose LDPET is extremely grainy and nearly unintelligible. While the individually trained U-Net successfully suppresses noise, it tends to "over-smooth" the image, creating a "waxy" appearance where fine anatomical details and edges are blurred—a clear manifestation of the "averaging effect." In contrast, our proposed method recovers sharper edges and preserves the contrast of internal structures more effectively, closely approximating the 100\% count ground truth. In the lower abdomen case (Fig. \ref{fig:bern_reginal}), although both models produce high-quality reconstructions at higher dose levels, our method consistently achieves superior quantitative performance across all metrics.

Fig. \ref{fig:bern_reginal_diff} presents the difference maps for UNN model and proposed methods at 2\% and 50\% dose levels. At the 50\% level, the denoised PET images from both methods closely approximate the FDPET reference. However, at the significantly lower 2\% dose level, our proposed method demonstrates superior stability, exhibiting fewer extreme intensity fluctuations (dark and bright artifacts) compared to the UNN approach. This improvement is even more pronounced at the 50\% level, where the difference map for our method is virtually negligible, indicating near-perfect reconstruction.

\subsubsection{A sample patient comparison}
To demonstrate the robustness of our proposed method across the entire body, we compared it against individually trained models at two distinct dose levels (2\% and 50\%) using the Shanghai Ruijin Hospital dataset. Figure \ref{fig:whole_body} presents these results using Maximum Intensity Projections (MIP), generated by selecting the maximum voxel intensity along the longitudinal (z) axis. We chose this visualization technique to provide a comprehensive view of the model's performance across all anatomical regions. The figure clearly illustrates that our method does not produce intensity outliers; in contrast, the individually trained U-Nets tend to overestimate intensities across most parts of the body, likely due to noise amplification. Ultimately, our method demonstrates consistent, high-quality performance at both the 2\% and 50\% dose levels.

\section{Discussion}
Generally, models trained directly on multi-dose data without prior dose information exhibit suboptimal performance, as demonstrated in the experiment section. In this work, we analyze the issue of models learning an 'averaged' LDPET representation and propose a streamlined approach using residual noise learning. By leveraging the LeakyReLU activation function, we enable the network to extract meaningful features from the noise while preventing dead neurons and invalid predictions. Notably, our method achieves superior results while maintaining a parameter count identical to that of individually trained dose-specific models.

In contrast to existing frameworks like UNN and ST-UNN, which rely on large-scale architectures and multiple sub-networks for individual noise levels, the proposed framework is substantially simpler to optimize and deploy. Our method does not require careful initialization or staged pretraining of unified subnetworks prior to joint optimization. Furthermore, our approach eliminates the need for weighted-sum aggregations, or the dual-model configurations required by UNN to achieve high performance. Our method also offers superior generalization; unlike the CADG approach, it does not rely on domain invariance, making it more robust.

As demonstrated in Table \ref{tab:comparison}, Fig. \ref{fig:std}, and Fig. \ref{fig:whole_body}, our method yields significantly more robust results. This stability aligns with intuitive deep learning principles; same as a classification model learns the shared essential features of a category, the "one-size-fits-all" extracts the fundamental anatomical structures common across various LDPET dose levels. While multi-dose training can significantly improve model robustness through exposure to data diversity, its success depends on avoiding the 'average learning' effect, which can otherwise severely degrade performance. By successfully mitigating this averaging pitfall while capitalizing on the strengths of multi-dose data, our model achieves a degree of generalization that directly enhances robustness \cite{liu2022personalized}.

Unlike explicit dose-conditioned methods, the proposed framework does not require external dose-level annotations during inference. Since the input LDPET image inherently encodes dose-dependent signal-to-noise characteristics, the network can partially exploit implicit conditioning through the observed degradation patterns. Reformulating the learning target into residual noise estimation further strengthens this dependency by emphasizing stochastic degradation components over anatomical reconstruction.

The proposed framework may be particularly relevant for real-world clinical PET deployment, where acquisition protocols, scanner hardware, injected dose, and reconstruction settings vary substantially across institutions. By improving robustness under heterogeneous count conditions without requiring explicit dose annotations, the proposed method may facilitate more practical cross-center PET denoising deployment.

Although the proposed framework focuses on optimization target reformulation rather than architectural scaling, further improvements may be achieved by integrating adaptive attention mechanisms or transformer-based feature modeling strategies, such as those employed in ST-UNN. In particular, combining residual noise learning with hierarchical self-attention may further improve cross-dose feature separation and robustness under highly heterogeneous acquisition conditions.

\section{Conclusion}
In summary, we have proposed a residual noise learning framework for cross-dose LDPET denoising. Our method effectively addresses the 'average learning' problem, enabling superior generalization and more robust performance across a wide range of noise levels. These findings demonstrate that our approach is a promising solution for practical applications involving variable dose inputs.

\section*{Acknowledgments}
Data used in the preparation of this article were obtained from the University of Bern, Department of Nuclear Medicine, and the School of Medicine, Ruijin Hospital. The investigators at these institutions contributed to the design and implementation of the data collection and/or provided the data but did not participate in the analysis or writing of this manuscript.


%\section{References Section}
%You can use a bibliography generated by BibTeX as a .bbl file.
% The IEEEtran BibTeX style support page is:
 
%\section{Simple References}
%You can manually copy in the resultant .bbl file and set second argument of $\backslash${\tt{begin}} to the number of references
% (used to reserve space for the reference number labels box).

\bibliographystyle{IEEEtran}
\bibliography{ref}

\newpage

% your own custom macro containing the $\backslash${\tt{includegraphics}} command to make things
 

%\bf{If you include a photo:}\vspace{-33pt}
%\begin{IEEEbiography}[{\includegraphics[width=1in,height=1.25in,clip,keepaspectratio]{fig1}}]{Michael Shell}
%Use $\backslash${\tt{begin\{IEEEbiography\}}} and then for the 1st argument use $\backslash${\tt{includegraphics}} to declare and link the author photo.
%Use the author name as the 3rd argument followed by the biography text.
%\end{IEEEbiography}



\vfill

\end{document}